\documentclass{article}
\usepackage{spconf,amsmath,graphicx}
\usepackage{multirow}
\usepackage{multicol}
\usepackage{booktabs}
\usepackage{graphicx}
\usepackage{boldline}
\usepackage[table, dvipsnames]{xcolor}
\usepackage{indentfirst}
\usepackage{graphicx,subcaption}
\usepackage[belowskip=-10pt,aboveskip=5pt]{caption}
\usepackage{caption} 
\title{n-RPN: hard example learning for region proposal networks}

\name{MyeongAh Cho, Tae-young Chung, Hyeongmin Lee and Sangyoun Lee\sthanks{Corresponding Author. \newline \hspace*{0.16in} This work was supported by Institute of Information \& communications Technology Planning \& Evaluation (IITP) grant funded by the Korea government(MSIT) (2016-0-00197, Development of the high-precision natural 3D view generation technology using smart-car multi sensors and deep learning)}}
\address{School of Electrical and Electronic Engineering, Yonsei University, Republic of Korea\\
	\{maycho0305, tato0220, minimonia, syleee\}@yonsei.ac.kr}

\begin{document}

\maketitle

\begin{abstract}

The region proposal task is to generate a set of candidate regions that contain an object. In this task, it is most important to propose as many candidates of ground-truth as possible in a fixed number of proposals. 
In a typical image, however, there are too few hard negative examples compared to the vast number of easy negatives, so region proposal networks struggle to train on hard negatives.
Because of this problem, networks tend to propose hard negatives as candidates, while failing to propose ground-truth candidates, which leads to poor performance. In this paper, we propose a Negative Region Proposal Network(nRPN) to improve Region Proposal Network(RPN). The nRPN learns from the RPN's false positives and provide hard negative examples to the RPN. Our proposed nRPN leads to a reduction in false positives and better RPN performance. An RPN trained with an nRPN achieves performance improvements on the PASCAL VOC 2007 dataset.
\end{abstract}


\begin{figure}[ht]
	\centering	
	\includegraphics[width=\columnwidth]{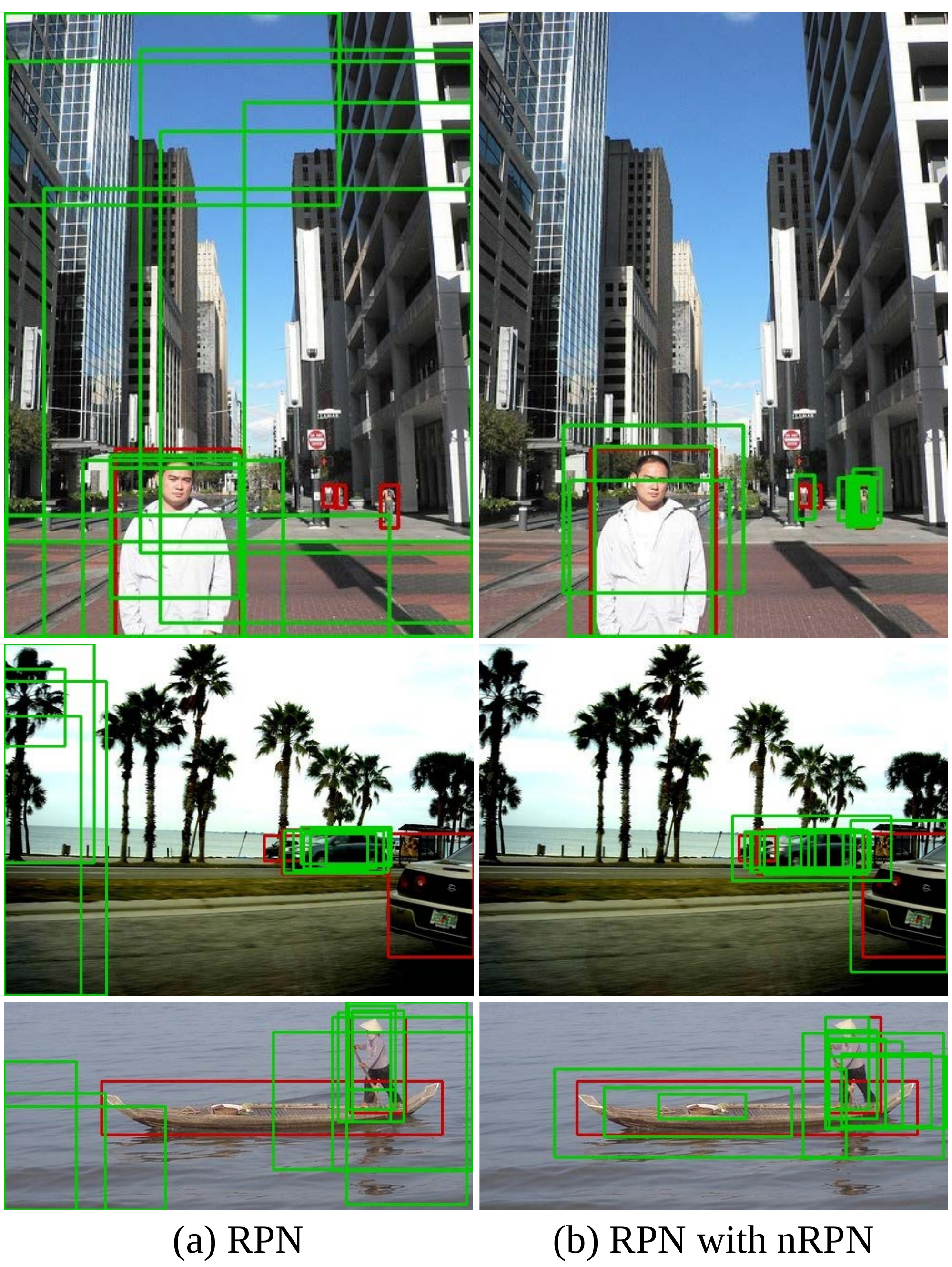}
	\caption{The result image of region proposal networks. The red boxes are ground-truth and green boxes are top 10 proposals of each image.}
	\label{result}
\end{figure}

\section{Introduction}
Region proposal networks are mainly used in the first stage of the region-based object detector by distinguishing objects from the background. In previous works, there are Selective Search method \cite{uijlings2013selective} which adapts segmentation with exhaustive search and Edge Boxes method \cite{zitnick2014edge} which generates bounding boxes from the edges. In Ren \textit{et al.} \cite{ren2015faster}, the Region Proposal Network(RPN) which predicts objectness score and coordinates of proposals with multiple scales and ratios of the anchors. Also, Lu \textit{et al.} \cite{ lu2016adaptive} proposes adaptive search strategy which recursively divides the image into sub-regions.
These methods bring the computational efficiency by reducing the number of candidate regions and better performance of a detector \cite{hosang2016makes}.

The region proposal networks learn objectness through a binary classification that classifies between two classes, foreground(positive) and background(negative). However, the foreground-background class imbalance is a challenging problem in the region proposal and object detection tasks. Compared to the foreground examples, the background examples are too easy which have low loss and it leads to degenerate the model. To solve this problem, we need hard negative examples which causes high losses. There are lots of efforts for hard negative mining such as Online Hard Example Mining(OHEM) \cite{shrivastava2016training} which train only high-loss proposals by selecting the examples that performs worst and \cite{ felzenszwalb2010object, zhang2016faster} which use boosted decision trees.
 
\begin{figure*}[ht]
	\centering
	\includegraphics[width=0.8\linewidth]{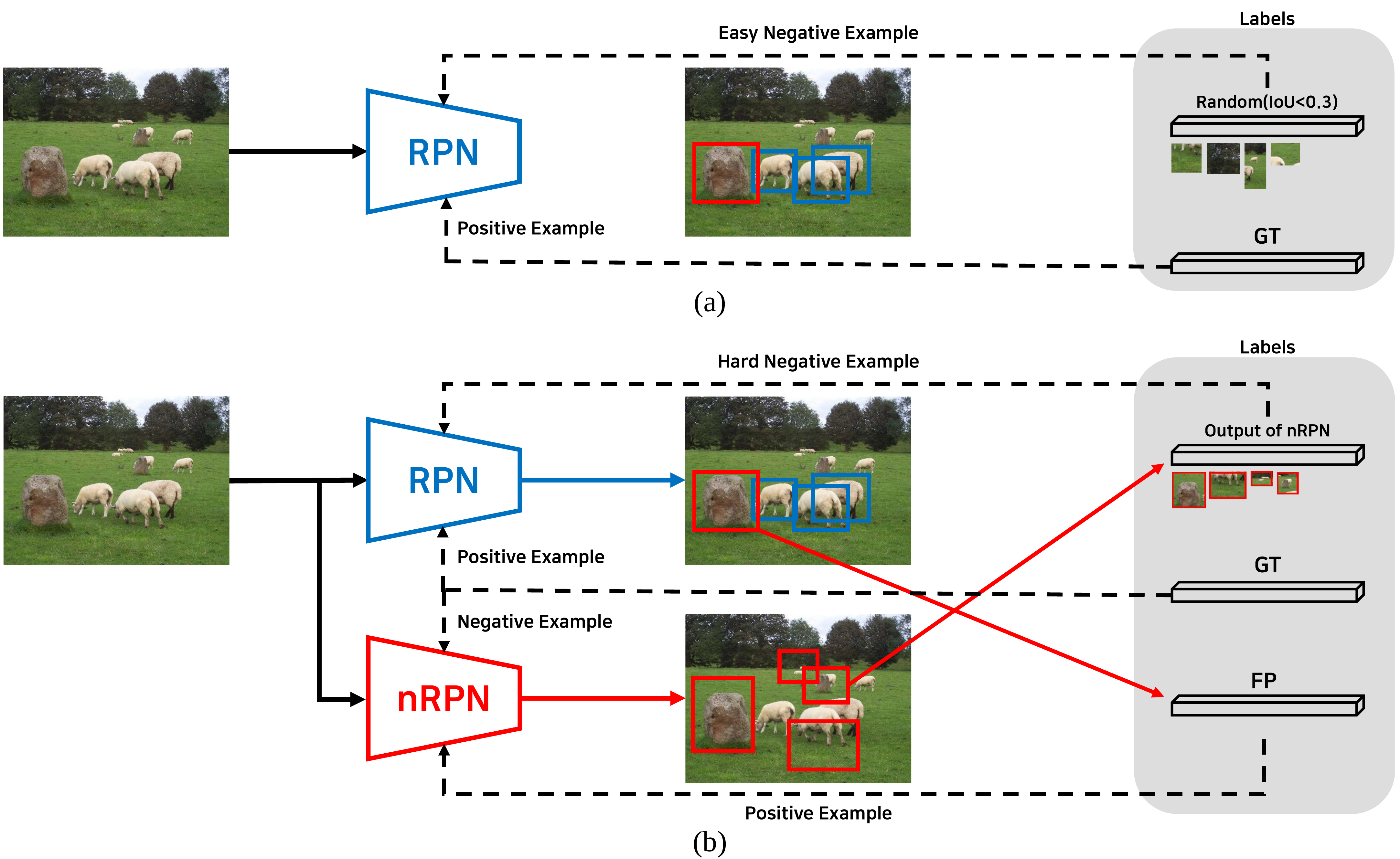}
	\caption{Framework of (a) original RPN and (b) RPN with nRPN  which both RPN and nRPN proposals contribute to each other's labels. }
	\label{framework}
\end{figure*}

In this paper, instead of hard example mining, we propose hard negative example learning network named Negative Region Proposal Network(nRPN). 
The nRPN aims to propose hard negative examples that the RPN might incorrect. nRPN trains with the false positives from RPN, in the meanwhile, RPN trains with the hard negative examples which are proposed by the nRPN. Both RPN and nRPN train at the same time, they provide positive or negative examples to each other and gradually generates more difficult examples. This approach leads higher recall of RPN and better performance of detector (Fig. \ref{result}). In addition, we propose loss function that considers the Intersection over Union (IoU) between each anchor and ground-truth(GT) to apply different loss according to IoU.
In this paper, our main contributions are,

\begin{itemize}
	\setlength{\itemsep}{0.5em}
	\setlength{\parskip}{0pt}
	\item Our proposed nRPN learns hard negative examples from false positives of RPN and provides hard negative examples to RPN. By training RPN and nRPN together, we can easily get hard negatives from nRPN which is only used for training. 
	\item Also we propose Overlap Loss to compute different loss according to overlap value of anchor and GT. The overlap loss is more effective for learning both the size of large and small objects.
\end{itemize}

\section{The proposed method}

\begin{figure*}[ht]
	\centering
	\includegraphics[scale=0.5]{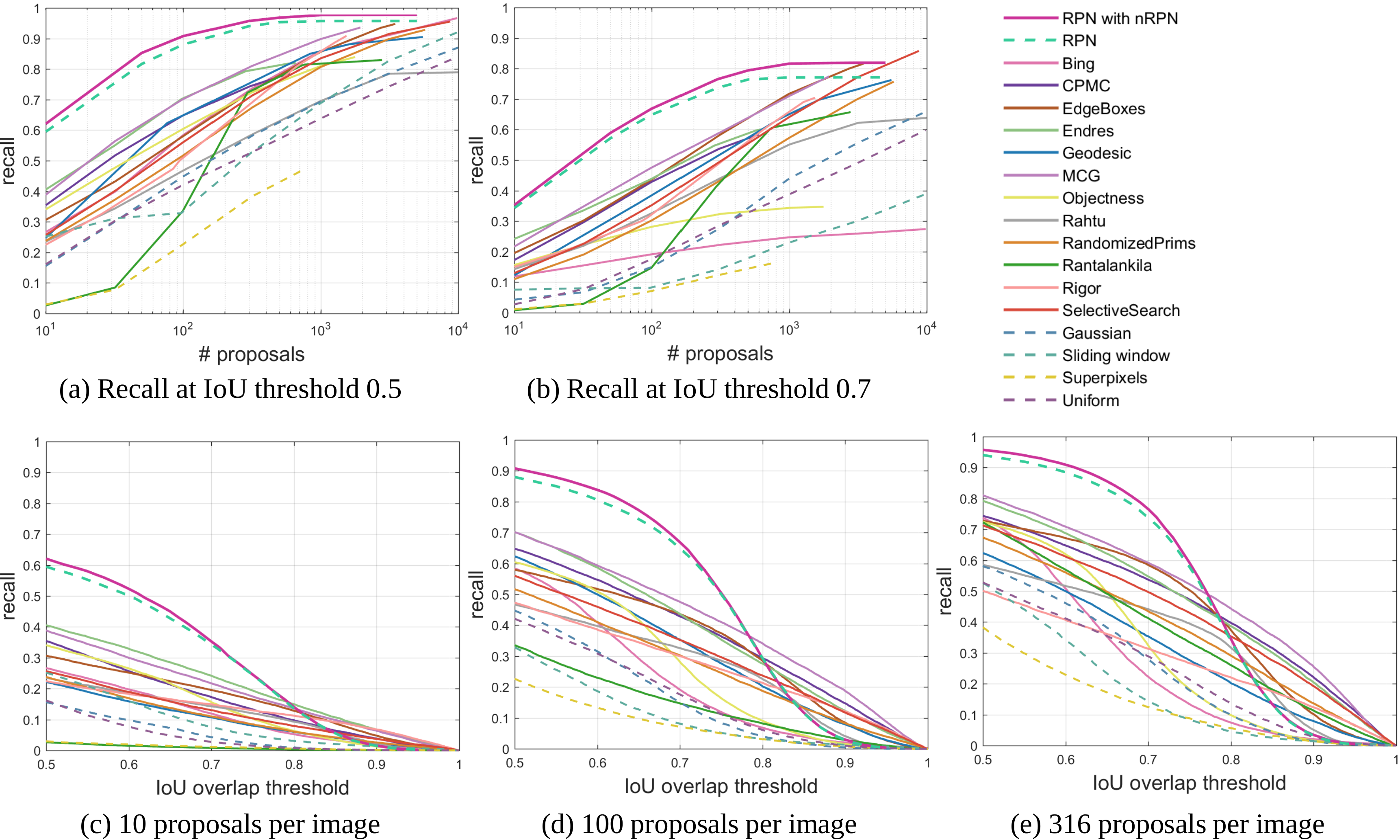}
	\caption{Recall rates of the proposed model and other region proposal models on PASCAL VOC 2007 testset.}
	\label{recall}
\end{figure*}

\subsection{Overview}

For the base region proposal model, we use RPN from faster-RCNN \cite{ren2015faster}. RPN performs region proposing through various scales and ratios of anchors which location is on each pixel of the feature map. RPN outputs the objectness score and coordinates of these anchors by passing input image through the feature extractor (VGG-16 \cite{simonyan2014very}) and sliding window.

Since Bengio \textit{et al.} \cite{bengio2009curriculum} shows that start learning with easier task and then gradually increasing the difficulty of learning can improve the generalization and faster convergence, we first train RPN without nRPN. In Fig. \ref{framework}(a), RPN label consists of GT and randomly selected easy negative examples which IoU with GT is lower than threshold($0.3$). After training a small number of epochs, the RPN learns objectness. Then, nRPN starts to train with false positives(FP) among the RPN proposals. As shown in Fig. \ref{framework} (b), proposals of nRPN is used as hard negative example for RPN and false positives of RPN goes positive examples for nRPN. Both networks are trained simultaneously but does not share weights each other. 

\subsection{nRPN}
nRPN is a network that learns hard negative examples and predicts false positives of RPN. Therefore, nRPN's positive examples are false positives which are incorrect example of the RPN output. Unlike detectors which false positives are very small amount \cite{jin2018unsupervised}, RPN has large number of false positives. Therefore, it’s easy to get false positives for nRPN. Also, since nRPN predicts the hard negative examples, RPN can easily trained with hard negatives than using other methods which mine hard examples and re-train with them. We defined false positive as an anchor which objectness score is higher than 0.7 but IoU with GT is lower than threshold($0.3$). On the other hand, the negative example of RPN is the proposal of nRPN excluding GT. Since nRPN is a network for predicting hard negative examples, it does not need bounding box regression.

\begin{figure}[t]
	\centering
	\includegraphics[width=\columnwidth]{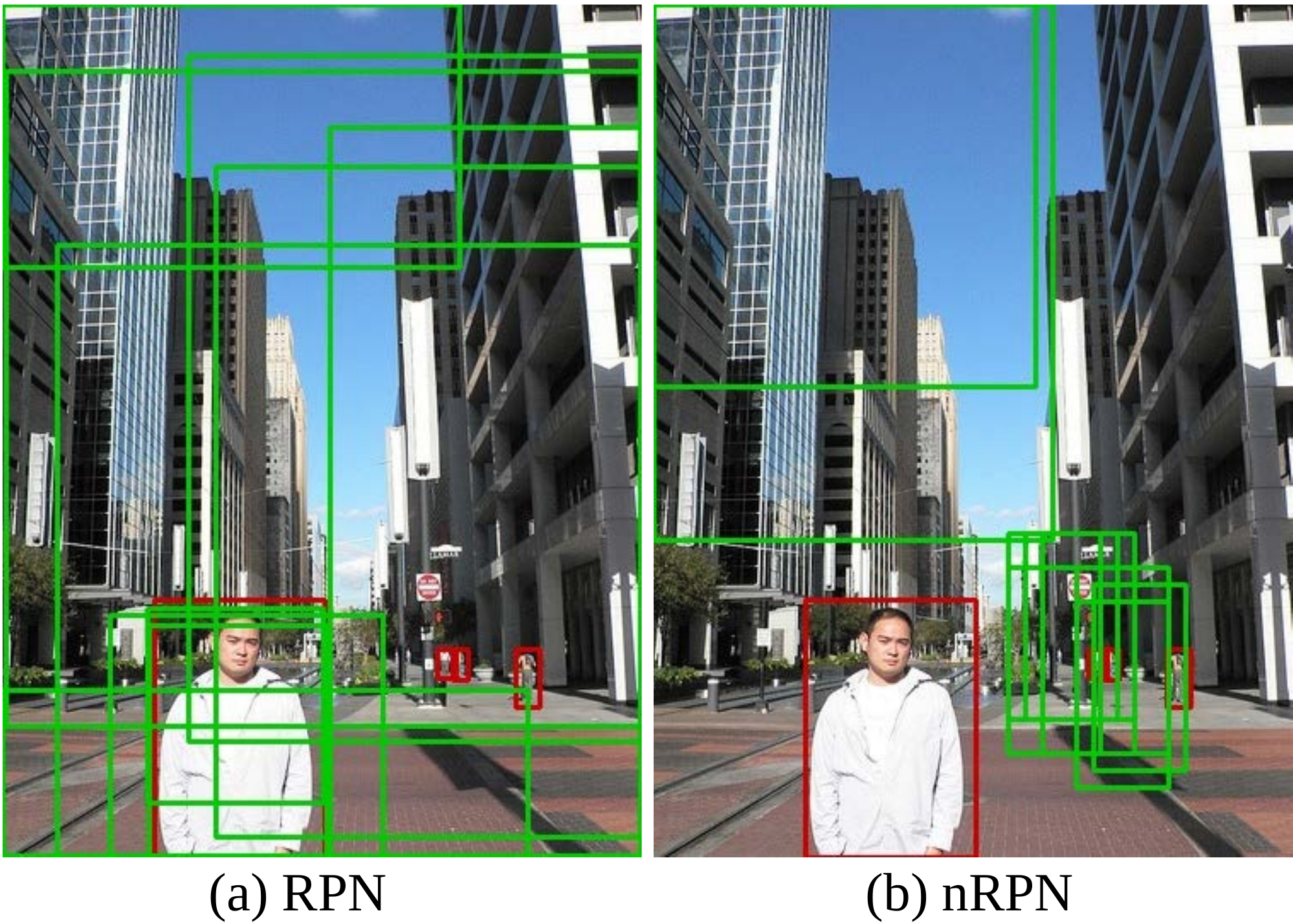}
	\caption{The region proposals of (a)RPN and (b)nRPN. The red boxes are ground-truth and the green boxes are proposals.}
	\label{nrpn}
\end{figure}

In Fig. \ref{nrpn}, we draw top 10 proposals of original RPN and nRPN. Since RPN propose lots of false positives with the high score, it fail to propose GT. In Fig. \ref{nrpn} (b), shows that proposals of nRPN are non-object region but RPN might wrong.

\begin{table*}[htbp]
	\centering
	\resizebox{\linewidth}{!}{
		\begin{tabular}{c|c|cccccccccccccccccccc|c}
			\toprule
			Model & \#Proposal & aero  & bicycle & bird  & boat  & bottle & bus   & car   & cat   & chair & cow   & table & dog   & horse & mbike & person & plant & sheep & sofa  & train & tv    & mAP(\%) \\
			\midrule
			\midrule
			\multirow{3}[2]{*}{RPN} & 50    & 67.1  & 74.02 & 60.41 & 49.72 & 38.5  & 74.37 & 78.73 & 80.81 & 42.3  & 74    & 63.92 & 76.54 & 82.57 & 72.15 & 69.81 & 38.71 & 65.23 & 62.66 & 74.29 & 57.68 & 65.18 \\
			& 100   & 69.12 & 77.78 & 64.7  & 51.78 & 44.52 & 75.92 & 79.4  & 84.41 & 42.13 & 73.56 & 66.5  & 77.75 & 82.34 & 71.98 & 75.87 & 39.64 & 69.02 & 63.6  & 76.09 & 64.39 & 67.52 \\
			& 300   & 69.97 & 78.56 & 65.86 & 54.16 & 48.49 & 78.85 & 82.99 & 84.6  & 42.73 & 75.94 & 66.97 & 80.08 & 82.85 & 74.48 & 76.31 & 40.89 & 69.64 & 63.68 & 75.31 & 65.95 & 68.92 \\
			\midrule
			\multicolumn{1}{c|}{\multirow{3}[2]{*}{\shortstack{RPN\\+nRPN}}} & 50    & 69.24 & 76.45 & 63.49 & 54.09 & 40.79 & 73.86 & 79.25 & 80.68 & 40.7  & 73.91 & 61.4  & 76.43 & 82.22 & 71.27 & 71.79 & 35.53 & 69.2  & 63.46 & 73.23 & 60.02 & 65.85 \\
			& 100   & 70.64 & 77.72 & 65.98 & 55.21 & 46.17 & 74.62 & 81.5  & 83.74 & 42.97 & 73.5  & 63.77 & 79.42 & 82.5  & 76.16 & 76.42 & 36.8  & 68.97 & 64.97 & 74.08 & 64.36 & 67.98 \\
			& 300   & 70.64 & 78.51 & 68.72 & 54.38 & 48.83 & 77.5  & 83.97 & 84.17 & 42.92 & 75.12 & 65.78 & 80.06 & 82.74 & 76.4  & 76.71 & 40.19 & 70.51 & 64.28 & 74.35 & 65.86 & 69.08 \\
			\midrule
			\multicolumn{1}{c|}{\multirow{3}[2]{*}{\shortstack{RPN\\+nRPN\\+$\mathit{L_{ov}}$}}} & 50    & 69.4  & 76.31 & 64.09 & 50.47 & 41.73 & 73.42 & 79.16 & 80.77 & 41.18 & 73.44 & 63.39 & 76    & 82.13 & 73.17 & 74.95 & 37.01 & 68.32 & 62.32 & 72.72 & 60.07 & \textbf{66} \\
			& 100   & 69.37 & 77.22 & 65.76 & 53.78 & 45.51 & 76.08 & 83.74 & 82.69 & 42.95 & 73.95 & 64.35 & 79.48 & 83.64 & 76.13 & 76.53 & 39.22 & 68.96 & 65.49 & 72.76 & 63.9  & \textbf{68.08} \\
			& 300   & 69.93 & 78.81 & 68.05 & 55.52 & 47.53 & 76.95 & 84    & 85.28 & 43.41 & 74.57 & 64.86 & 79.21 & 83.33 & 77.02 & 76.85 & 41.57 & 70.11 & 65.62 & 73.62 & 66.17 & \textbf{69.12} \\
			\bottomrule
	\end{tabular}}
	\caption{mean Average Precision(\%) of RPN and proposed approach with Faster-RCNN on PASCAL VOC 2007 dataset.}
	\label{t2}
\end{table*}

\subsection{Overlap Loss}
As we mentioned above, each anchors are labeled with the foreground $ p^{*} = 1$ and the background $ p^{*} = 0$. Each foreground anchors have different IoU value with GT, which means the probability of being an object is also different. That is, the expected objectness score for each foreground anchor should be considered its IoU value rather than 1. Therefore, according to IoU between GT and each foreground anchor, the predicted objectness score $p_{i}$ is divided by IoU in Eq (\ref{eq1}). 

\begin{gather}
{p_{i}}' = \begin{cases}
\frac{p_{i}}{IoU}& \text{ if } p_{i}^{*}= 1\\ 
1 - p_{i}& \text{ if } p_{i}^{*}= 0
\end{cases} \nonumber\\
L_{ov}\left ( p_{i}, p_{i}^{*} \right ) = -p_{i}^{*}\, log \, {p_{i}}' - \left ( 1-p_{i}^{*} \right )\, log\left ( 1-{p_{i}}' \right )
\label{eq1}
\end{gather}
\begin{gather}
L = \frac{1}{N_{cls}}\sum_{i}L_{ov}\left ( p_{i}, p_{i}^{*} \right ) + \lambda \frac{1}{N_{reg}}\sum_{i}L_{reg}\left ( t_{i}, t_{i}^{*} \right )
\label{eq2}
\end{gather}

Since small objects tends to have a lower IoU with anchor than large objects, it is hard to train in RPN. However, this overlap loss can help to learn more balanced with object size. In Eq (\ref{eq1}) $p_{i}^{*}$ and $p_{i}$ denotes GT label and predicted probability of anchor $i$. We call this loss as overlap loss $L_{ov}$. The total loss $L$ is in Eq (\ref{eq2}) where $L_{reg}$ is smooth $L1$. 

\begin{table}[h]
	\centering
	\resizebox{\columnwidth}{!}{
		\begin{tabular}{c|c|ccc|ccc|ccc}
			\toprule
			\multicolumn{2}{c|}{\multirow{2}[4]{*}{Recall(\%)}} & \multicolumn{3}{c|}{RPN} & \multicolumn{3}{c|}{RPN+nRPN} & \multicolumn{3}{c}{RPN+nRPN+$\mathit{L_{ov}}$} \\
			\cmidrule{3-11}    \multicolumn{2}{c|}{} & 0.5   & 0.7   & 0.9   & 0.5   & 0.7   & 0.9   & 0.5   & 0.7   & 0.9 \\
			\midrule
			\midrule
			\multirow{4}[2]{*}{50} & S     & 48.7  & 21.0  & 0.4   & 55.2  & \textbf{28.4} & \textbf{0.7} & \textbf{56.4} & 27.7  & 0.4 \\
			& M     & 66.7  & 39.6  & 1.1   & 68.9  & 43.6  & 1.7   & \textbf{73.2} & \textbf{46.6} & \textbf{2.0} \\
			& L     & 91.6  & 68.6  & 3.4   & 93.5  & \textbf{68.0} & \textbf{3.0}   & \textbf{93.5} & 67.4  & \textbf{3.0} \\
			\rowcolor{lightgray!70}
			& all   & 81.7  & 57.1  & 2.5   & 83.9  & 58.4  & 2.5   & \textbf{85.3} & \textbf{58.9} & \textbf{2.6} \\
			\midrule
			\multirow{4}[2]{*}{100} & S     & 54.7  & 23.8  & 0.4   & 60.1  & \textbf{31.2} & 0.7   & \textbf{61.4} & 30.0  & \textbf{0.7} \\
			& M     & 77.8  & 48.4  & 0.3   & 80.0  & 52.5  & 2.1   & \textbf{83.0} & \textbf{55.7} & \textbf{2.3} \\
			& L     & 95.6  & 76.2  & 3.9   & \textbf{97.1} & \textbf{76.8} & 3.3   & 96.9  & 75.2  & \textbf{3.5} \\
			\rowcolor{lightgray!70}
			& all   & 88.0  & 64.9  & 2.9   & 89.9  & \textbf{66.9} & 2.8   & \textbf{90.8} & \textbf{66.9}  & \textbf{3.0} \\
			\midrule
			\multirow{4}[2]{*}{300} & S     & 65.6  & 27.5  & 0.5   & 68.4  & \textbf{35.1} & 0.9   & \textbf{69.1} & 33.0  & \textbf{0.9} \\
			& M     & 89.1  & 59.7  & 1.7   & 91.2  & 65.3  & 2.6   & \textbf{92.9} & \textbf{67.5} & \textbf{2.7} \\
			& L     & 98.7  & 84.1  & 4.2   & \textbf{99.3} & \textbf{85.7} & 3.9   & 99.2  & 84.3  & \textbf{4.0} \\
			\rowcolor{lightgray!70}
			& all   & 94.1  & 73.7  & 3.2   & 95.2  & \textbf{76.8} & 3.3   & \textbf{95.7} & 76.5  & \textbf{3.4} \\
			\bottomrule
	\end{tabular}}
	\caption{Results of the RPN and proposed approach on the PASCAL VOC 2007 dataset.}
	\label{t1}
\end{table}

\section{Experiment}

For experiment, we use PASCAL VOC 2007 \cite{everingham2010pascal} trainval and test dataset. It consists of about 5,000 images each of the trainval and testset over 20 object categories. Our base model is RPN with VGG-16 which is pre-trained on ImageNet. For nRPN, we use same structure with RPN which is consists of 13 convolutional layers for feature extractor and 2 convolutional layers for score map. The batch size is 1 and all the models are trained for 20 epochs.

\subsection{Results of Proposed Method }
In Table \ref{t1}, we compared the RPN results and our proposed models. It shows the recall (\%) with IoU threshold of 0.5, 0.7 and 0.9 when each proposal number is 50, 100, and 300. We also compute recalls when object size is small($\alpha<32^{2}$), medium($32^{2}\leq \alpha<96^{2}$) and large($96^{2}\leq\alpha$) where $\alpha$ is area of an object. Compare to the RPN, the model trained with hard negatives which is proposed by nRPN performed better. Based on the IoU threshold of 0.7, it shows 1.3\%, 2\%, and 3.1\% improvements in the 50, 100, and 300 number of proposals. In addition, model that trained using overlap loss also improved performance on small, medium size of objects. It means that overlap loss that consider IoU of foreground anchor helps to train properly according to the size of an object.

We plot recall rates of RPN trained with nRPN model and other region proposal models according to number of proposals and IoU threshold in Fig. \ref{recall}. For comparing models, we use Bing \cite{cheng2014bing}, CPMC \cite{carreira2010constrained}, EdgeBoxes \cite{zitnick2014edge}, Endres \cite{endres2010category}, Geodesic \cite{krahenbuhl2014geodesic}, MCG \cite{arbelaez2014multiscale}, Objectness \cite{alexe2010object}, Rahtu \cite{rahtu2011learning}, RandomizedPrims \cite{hosang2016makes}, Rantalankila \cite{rantalankila2014generating}, Rigor \cite{humayun2014rigor}, SelectiveSearch \cite{uijlings2013selective}, Gaussian, SlidingWindow, Superpixels, Uiform \cite{hosang2016makes}, RPN \cite{ren2015faster}. 

In the first row of Fig. \ref{recall}, training RPN with our proposed nRPN and overlap loss shows performance improvement over the original RPN. Also in the Fig. \ref{recall} (d), RPN with nRPN performed over 90\% recall at IoU threshold 0.5. The RPN with nRPN shows mostly better performance than others.

\subsection{Precision of Object Detection with nRPN}
We apply RPN and our proposed model to the faster-RCNN\cite{ren2015faster} object detection network. Table \ref{t2} shows mean Average Precision(mAP) and Average Precision(AP) of each category in PASCAL VOC 2007 dataset at 50, 100 and 300 number of proposals from region proposal network. The RPN trained with nRPN and overlap loss shows improvements on detector precision which means that the better performance of RPN leads improvement of the detector performance.

\section{Conclusion}
\label{sec:Conclusion}

In this paper, we propose nRPN which is the hard negative example learning network for region proposal task. Instead of hard example mining method, this simple nRPN network train with false positives of RPN and provide hard negatives to improve RPN during the training time. Also, we suggest overlap loss to learn more balanced with object size. This proposed loss helps to train more properly with small-medium size objects in anchor based detector. We showed that our proposed approach improved RPN performance and detector precision on PASCAL VOC 2007.

\vfill
\pagebreak

\bibliographystyle{IEEEbib}
\bibliography{strings,refs}

\begin{thebibliography}{10}

\bibitem{uijlings2013selective}
Jasper~RR Uijlings, Koen~EA Van De~Sande, Theo Gevers, and Arnold~WM Smeulders,
\newblock ``Selective search for object recognition,''
\newblock {\em International journal of computer vision}, vol. 104, no. 2, pp.
  154--171, 2013.

\bibitem{zitnick2014edge}
C~Lawrence Zitnick and Piotr Doll{\'a}r,
\newblock ``Edge boxes: Locating object proposals from edges,''
\newblock in {\em European conference on computer vision}. Springer, 2014, pp.
  391--405.

\bibitem{ren2015faster}
Shaoqing Ren, Kaiming He, Ross Girshick, and Jian Sun,
\newblock ``Faster r-cnn: Towards real-time object detection with region
  proposal networks,''
\newblock in {\em Advances in neural information processing systems}, 2015, pp.
  91--99.

\bibitem{lu2016adaptive}
Yongxi Lu, Tara Javidi, and Svetlana Lazebnik,
\newblock ``Adaptive object detection using adjacency and zoom prediction,''
\newblock in {\em Proceedings of the IEEE Conference on Computer Vision and
  Pattern Recognition}, 2016, pp. 2351--2359.

\bibitem{tychsen2017denet}
Lachlan Tychsen-Smith and Lars Petersson,
\newblock ``Denet: Scalable real-time object detection with directed sparse
  sampling,''
\newblock {\em arXiv preprint arXiv:1703.10295}, 2017.

\bibitem{hosang2016makes}
Jan Hosang, Rodrigo Benenson, Piotr Doll{\'a}r, and Bernt Schiele,
\newblock ``What makes for effective detection proposals?,''
\newblock {\em IEEE transactions on pattern analysis and machine intelligence},
  vol. 38, no. 4, pp. 814--830, 2016.

\bibitem{shrivastava2016training}
Abhinav Shrivastava, Abhinav Gupta, and Ross Girshick,
\newblock ``Training region-based object detectors with online hard example
  mining,''
\newblock in {\em Proceedings of the IEEE Conference on Computer Vision and
  Pattern Recognition}, 2016, pp. 761--769.

\bibitem{felzenszwalb2010object}
Pedro~F Felzenszwalb, Ross~B Girshick, David McAllester, and Deva Ramanan,
\newblock ``Object detection with discriminatively trained part-based models,''
\newblock {\em IEEE transactions on pattern analysis and machine intelligence},
  vol. 32, no. 9, pp. 1627--1645, 2010.

\bibitem{zhang2016faster}
Liliang Zhang, Liang Lin, Xiaodan Liang, and Kaiming He,
\newblock ``Is faster r-cnn doing well for pedestrian detection?,''
\newblock in {\em European Conference on Computer Vision}. Springer, 2016, pp.
  443--457.

\bibitem{simonyan2014very}
Karen Simonyan and Andrew Zisserman,
\newblock ``Very deep convolutional networks for large-scale image
  recognition,''
\newblock {\em arXiv preprint arXiv:1409.1556}, 2014.

\bibitem{bengio2009curriculum}
Yoshua Bengio, J{\'e}r{\^o}me Louradour, Ronan Collobert, and Jason Weston,
\newblock ``Curriculum learning,''
\newblock in {\em Proceedings of the 26th annual international conference on
  machine learning}. ACM, 2009, pp. 41--48.

\bibitem{jin2018unsupervised}
SouYoung Jin, Aruni RoyChowdhury, Huaizu Jiang, Ashish Singh, Aditya Prasad,
  Deep Chakraborty, and Erik Learned-Miller,
\newblock ``Unsupervised hard example mining from videos for improved object
  detection,''
\newblock in {\em Proceedings of the European Conference on Computer Vision
  (ECCV)}, 2018, pp. 307--324.

\bibitem{everingham2010pascal}
Mark Everingham, Luc Van~Gool, Christopher~KI Williams, John Winn, and Andrew
  Zisserman,
\newblock ``The pascal visual object classes (voc) challenge,''
\newblock {\em International journal of computer vision}, vol. 88, no. 2, pp.
  303--338, 2010.

\bibitem{cheng2014bing}
Ming-Ming Cheng, Ziming Zhang, Wen-Yan Lin, and Philip Torr,
\newblock ``Bing: Binarized normed gradients for objectness estimation at
  300fps,''
\newblock in {\em Proceedings of the IEEE conference on computer vision and
  pattern recognition}, 2014, pp. 3286--3293.

\bibitem{carreira2010constrained}
Joao Carreira and Cristian Sminchisescu,
\newblock ``Constrained parametric min-cuts for automatic object
  segmentation,''
\newblock in {\em Computer Vision and Pattern Recognition (CVPR), 2010 IEEE
  Conference on}. IEEE, 2010, pp. 3241--3248.

\bibitem{endres2010category}
Ian Endres and Derek Hoiem,
\newblock ``Category independent object proposals,''
\newblock in {\em European Conference on Computer Vision}. Springer, 2010, pp.
  575--588.

\bibitem{krahenbuhl2014geodesic}
Philipp Kr{\"a}henb{\"u}hl and Vladlen Koltun,
\newblock ``Geodesic object proposals,''
\newblock in {\em European conference on computer vision}. Springer, 2014, pp.
  725--739.

\bibitem{arbelaez2014multiscale}
Pablo Arbel{\'a}ez, Jordi Pont-Tuset, Jonathan~T Barron, Ferran Marques, and
  Jitendra Malik,
\newblock ``Multiscale combinatorial grouping,''
\newblock in {\em Proceedings of the IEEE conference on computer vision and
  pattern recognition}, 2014, pp. 328--335.

\bibitem{alexe2010object}
Bogdan Alexe, Thomas Deselaers, and Vittorio Ferrari,
\newblock ``What is an object?,''
\newblock in {\em Computer Vision and Pattern Recognition (CVPR), 2010 IEEE
  Conference on}. IEEE, 2010, pp. 73--80.

\bibitem{rahtu2011learning}
Esa Rahtu, Juho Kannala, and Matthew Blaschko,
\newblock ``Learning a category independent object detection cascade,''
\newblock in {\em Computer Vision (ICCV), 2011 IEEE International Conference
  on}. IEEE, 2011, pp. 1052--1059.

\bibitem{rantalankila2014generating}
Pekka Rantalankila, Juho Kannala, and Esa Rahtu,
\newblock ``Generating object segmentation proposals using global and local
  search,''
\newblock in {\em Proceedings of the IEEE conference on computer vision and
  pattern recognition}, 2014, pp. 2417--2424.

\bibitem{humayun2014rigor}
Ahmad Humayun, Fuxin Li, and James~M Rehg,
\newblock ``Rigor: Reusing inference in graph cuts for generating object
  regions,''
\newblock in {\em Proceedings of the IEEE Conference on Computer Vision and
  Pattern Recognition}, 2014, pp. 336--343.

\end{thebibliography}

\end{document}